\documentclass[letterpaper, 10 pt, conference]{ieeeconf}  
\IEEEoverridecommandlockouts
\usepackage{cite}
\usepackage{amsmath,amssymb,amsfonts}
\usepackage{algorithmic}
\usepackage{graphicx}
\usepackage{textcomp}
\usepackage{xcolor}
\def\BibTeX{{\rm B\kern-.05em{\sc i\kern-.025em b}\kern-.08em
    T\kern-.1667em\lower.7ex\hbox{E}\kern-.125emX}}
\usepackage{float}
\usepackage[normalem]{ulem}
\usepackage{multirow}
\usepackage{soul}
\usepackage{tabularray}
\usepackage{caption}

\usepackage{amsmath,amsfonts}
\usepackage{algorithmic}
\usepackage{array}
\usepackage[caption=false,font=normalsize,labelfont=sf,textfont=sf]{subfig}
\usepackage{textcomp}
\usepackage{stfloats}

\usepackage{tabularray}
\UseTblrLibrary{diagbox}
\usepackage{url}
\usepackage{verbatim}
\usepackage{xcolor}
\usepackage{graphicx}
\usepackage{graphics}
\usepackage{algorithm2e}
\usepackage{adjustbox}
\usepackage{hyperref}
\usepackage{amsmath,amssymb,lipsum}

\usepackage{cuted} 
\usepackage{booktabs}
\usepackage{balance}
\usepackage[normalem]{ulem}
\usepackage[utf8]{inputenc}
\usepackage{cite}
\usepackage{lipsum, babel}
\begin{document}
\pagenumbering{arabic}
\bibliographystyle{IEEEtran}
\newcommand{\RNum}[1]{\lowecase\expandafter{\romannumeral #1\relax}}
\title{Counterfactual Explainer Framework for Deep Reinforcement Learning Models Using Policy Distillation}
\author{{Amir Samadi}$^{1}$, Konstantinos Koufos$^{1}$, Kurt Debattista$^{1}$ and Mehrdad Dianati$^{1,2},\emph{Senior Member, IEEE}$
\thanks{$^{1}$The authors are with the Warwick Manufacturing Group (WMG),
The University of Warwick, Coventry CV4 7AL. (e-mail:amir.samadi, konstantinos.koufos, k.debattista and m.dianati@warwick.ac.uk). This research is sponsored by Centre for Doctoral Training to Advance the Deployment of Future Mobility Technologies
(CDT FMT) at the University of Warwick.}
\thanks{$^{2}$The author is also with the School of Electronics, Electrical Engineering and Computer Science (EEECS), Queen’s University of Belfast (e-mail: m.dianati@qub.ac.uk).}}


\maketitle
\thispagestyle{plain}
\pagestyle{plain}

\begin{abstract} Deep Reinforcement Learning (DRL) has demonstrated promising capability in solving complex control problems. However, DRL applications in safety-critical systems are hindered by the inherent lack of robust verification techniques to assure their performance in such applications. One of the key requirements of the verification process is the development of effective techniques to explain the system functionality, providing why the system produces specific results in given circumstances. Recently, interpretation methods based on the Counterfactual (CF) explanation approach have been proposed to address the problem of explanation in DRLs. This paper proposes a novel CF explanation framework to interpret the decisions made by a black-box DRL. To evaluate the efficacy of the proposed explanation framework, we carried out several experiments in the domains of automated driving systems and Atari Pong game. Our analysis demonstrates that the proposed framework generates plausible and meaningful explanations for various decisions made by deep underlying DRLs. Source codes are available at: \url{https://github.com/Amir-Samadi/Counterfactual-Explanation}.
\end{abstract}


\section{Introduction}
Recent breakthroughs in solving various challenging problems by Deep Reinforcement Learning (DRL) models have been demonstrated in the literature~\cite{silver2016mastering,mnih2015human}. However, deploying DRL models to safety-critical applications, such as automated driving systems~\cite{tampuu2020survey} and healthcare \cite{mahmud2018applications}, is impeded due to lack of ability to ensure the trustworthiness of such models~\cite{tjoa2020survey}. To address this challenge, the concept of Interpretable Artificial Intelligence (IAI) has recently emerged as a promising solution~\cite{bacciu2022explaining,chou2022counterfactuals}. Within the realm of IAI, interpretability refers to the endeavour of generating human-understandable explanations for the functionality of black-box AI systems, benefiting various stakeholders, including end-users, engineers and legal authorities. To this end, in this paper, we investigate the use of Counterfactual (CF) reasoning, a promising IAI paradigm, for explaining the behaviour of DRL-based models particularly in the context of Automated Driving System~(ADS). 



The literature offers various CF reasoning methodologies to interpret the decisions made by black-box Deep Learning (DL) models. Pioneering work, Wachter et al.~\cite{wachter2017counterfactual} introduced CF explanation as the minimum changes required in the input of the black-box model to alter a given output to its complement, highlighting the most crucial features in the original input-output pairs. In principle, finding such necessary and minimum changes shapes the CF explanation as an optimisation problem. 
A range of optimisation methods have been proposed for CF explanations, including gradient-based optimisation (WachterCF)~\cite{wachter2017counterfactual}, genetic algorithms~\cite{guidotti2018local, sharma2019certifai, dandl2020multi}, game theory~\cite{ramon2020comparison, rathi2019generating} and monte-carlo methods~\cite{lucic2020does}. While these techniques have demonstrated promise in low-dimensional-input systems, such as tabular data, they are ineffective in systems with high-dimensional inputs, as in computer vision applications where high-resolution images must be processed~\cite{steex}(see section~\ref{subsection:counterfactual_explanation} for further discussion). 

To overcome this challenge, deep generative CF methods \cite{steex, rodriguez2021beyond, shrikumar2017learning, samadi20203SAFE} have been proposed, which generate CF explanations directly from specified input attributes using Generative Adversarial Networks (GANs). However, it's worth noting that these methods cannot guarantee the plausibility and meaningfulness of the generated CFs; This stems from the fact that GANs generate synthetic data rather than modifying the original data specifications, such as the localisation of Participant Vehicles~(PVs) in the context of ADSs~\cite{olson2021counterfactual}.
The term ``plausibility'' underscores the necessity of generated CF examples being both legitimate and actionable. For instance, in ADS, a fundamental rule dictates that all PVs must remain on the appropriate side of the road, making a CF state with an off-road vehicle implausible. On the other hand, ``meaningfulness'' implies that any alternation should be reasonable and sensible. For example, a minor change in the color of a leading PV is not comprehensible.

Generating high-dimensional CF examples poses a greater challenge in the realm of DRL agents due to the temporal dependencies introduced by processing a history length of observations at every time step~\cite{huber2023ganterfactual}. For instance, consider a DRL-based ADS which receives the latest four stacked states observations of the driving environment in each time step to provide an action. To generate a CF explanation for such a DRL agent, firstly, four CF states must be provided, which results in multiplying the high-dimensional input by the history observation length. Secondly, all four generated CF states should demonstrate internal consistency to represent temporal dependencies accurately. This means that in a highway driving scenario, the locations of vehicles in subsequent CF state images must mirror realistic driving manoeuvres, showcasing spatiotemporal correlations while adhering to speed limits and road geometry. Due to these challenges, to the best of our knowledge, there are currently only two studies in the literature designed to generate CF examples for a given DRL agent handling high-dimensional visual data. The first, by Olson et al.~\cite{olson2021counterfactual}, relies on an intricate combination of deep generative models to provide CF states for altering the DRL's action. However, the user cannot specify the desirable alternate action of the DRL agent, which limits its applicability. Besides, Olson's model fails to generate CF state in complex reinforcement learning environments such as ADS, proved in section~\ref{section:experiments}. The second model, developed by Huber et al.~\cite{huber2023ganterfactual} and known as GANterfactual-RL, leverages a well-establised image-to-image translation model, namely StarGAN~\cite{choi2018stargan}, to transfer the original state-action pairs to CF states associated with complement actions. Despite shedding some light on the behaviour of the DRL agents, both models often produce CFs that are implausible or lack meaningful interpretation~(see sections \ref{subsection:counterfactual_explanation} and \ref{section:experiments}). This limitation stems from the reliance of the utilized deep generative models solely on the observed states, lacking further direction on how to narrow down the solution space. This can result in suboptimal CF solutions, especially within the expansive visual domain. For instance, they fail to consider the inherent environmental limitations that would filter out impractical CF states, like generating vehicles off-road. Consequently, the existing literature lacks a proficient model capable of generating CF examples that are plausible and offer meaningful insights for DRL agents processing visual input states.

The work in this paper extends the existing literature by proposing an explanation framework (hereinafter called PDCF) that utilises policy distillation to address the challenge of generating meaningful and plausible CFs for DRL-based models with high-dimensional input. The proposed PDCF framework comprises (i) training a DRL-based control system with a high-dimensional input state, such as bird's eye view image of a highway, hereafter referred to as the teacher model, (ii) distilling the teacher's learned policies to a student network with a low-dimensional input in the form of human-understandable features, such as the objectives list of the PVs in the highway (iii) providing candidate CFs for the student network, and finally (iv) verifying whether the teacher and student networks provide similar outputs for the candidate CFs. This approach alleviates the need to solve optimisation problems in high-dimensional input spaces. Furthermore, finding the CF explanations by altering human-understandable low-dimensional features, such as the positions of vehicles (belonging to the objective list attributes), ensures the meaningfulness and plausibility of the generated CFs. To the best of our knowledge, the proposed explanation framework is the first method that guarantees the plausibility of generated CF explanations (see section \ref{subsection:counterfactual_explanation}).

The main contributions of this paper are also summarised in the list below:
\begin{enumerate}
    \item A novel CF explainer framework named PDCF is proposed, which can be applied to any DRL model, using policy distillation.
    \item We discuss how stakeholders, such as end-users and engineers can utilise the CF explanations. In particular, using the proposed explanation framework, we extract human-understandable if-then rules from black-box DRL models to explain the learned policies.
    \item A comprehensive analysis of the proposed framework is carried out in the context of three DRL-based systems, namely, the deep Q-network (DQN)~\cite{mnih2015human}, the asynchronous actor-critic (A2C) algorithm~\cite{mnih2016asynchronous} and the proximal policy optimization (PPO)~\cite{schulman2017proximal} acting in Atari Pong game \cite{1606.01540}, highway and roundabout driving \cite{highway-env}.
\end{enumerate}

The rest of this paper is organised as follows. We summarise the related works in the field of counterfactual explanation and review the policy distillation method in Section~\ref{section:Related_work}. Section~\ref{section: method} provides technical aspects of the proposed method. The conducted experiments, further discussion and research findings are described in Sections~\ref{section:experiments} and \ref{sec: Discussion} and \ref{section: conlusion}, respectively.

\section{Background and Related Work}
\label{section:Related_work}
This section presents a brief overview of the background and related work in the realm of interpretability, focusing on the use of counterfactual (CF) explanations as a promising approach to achieve interpretability. In addition, we review the policy distillation method as the key technique employed in this study, which simplifies Deep Reinforcement Learning (DRL) models, thereby enabling the generation of CF explanation. 

\subsection{Interpretability}
Intending to open the AI black-box, we have recently witnessed considerable research on interpretability methods. The family of interpretable models can be divided into two main streams: Intrinsic and post-hoc techniques. While intrinsic interpretable algorithms refer to techniques that could generate explanations during the training phase~\cite{kim2019grounding}, post-hoc methods employ auxiliary models to extract an explanation from already trained models~\cite{bojarski2016end}. Intrinsic models are also known as transparent models, implying that several models belonging to this category, such as decision trees, linear models, logistic models, and the k-nearest neighbours algorithm, are intrinsically interpretable due to their straightforward structure.

The post-hoc interpretable algorithms can be categorised as model-specific or model-agnostic \cite{chou2022counterfactuals}. Model-specific interpretability approaches consider the model as a "grey box" in which model internals, such as weights and features, are accessible, and explanations can be produced around these features and weights. On the other hand, model-agnostic methods treat the model as a "black-box" whose input and output can only be probed. In this manner, a surrogate model is widely used to learn a behaviour similar to that of the black-box model, and this surrogate (or proxy) model is then interpreted. The CF paradigms, including our study, are mainly classified as model-agnostic.

\begin{figure}
\centering
\includegraphics[width=\linewidth]{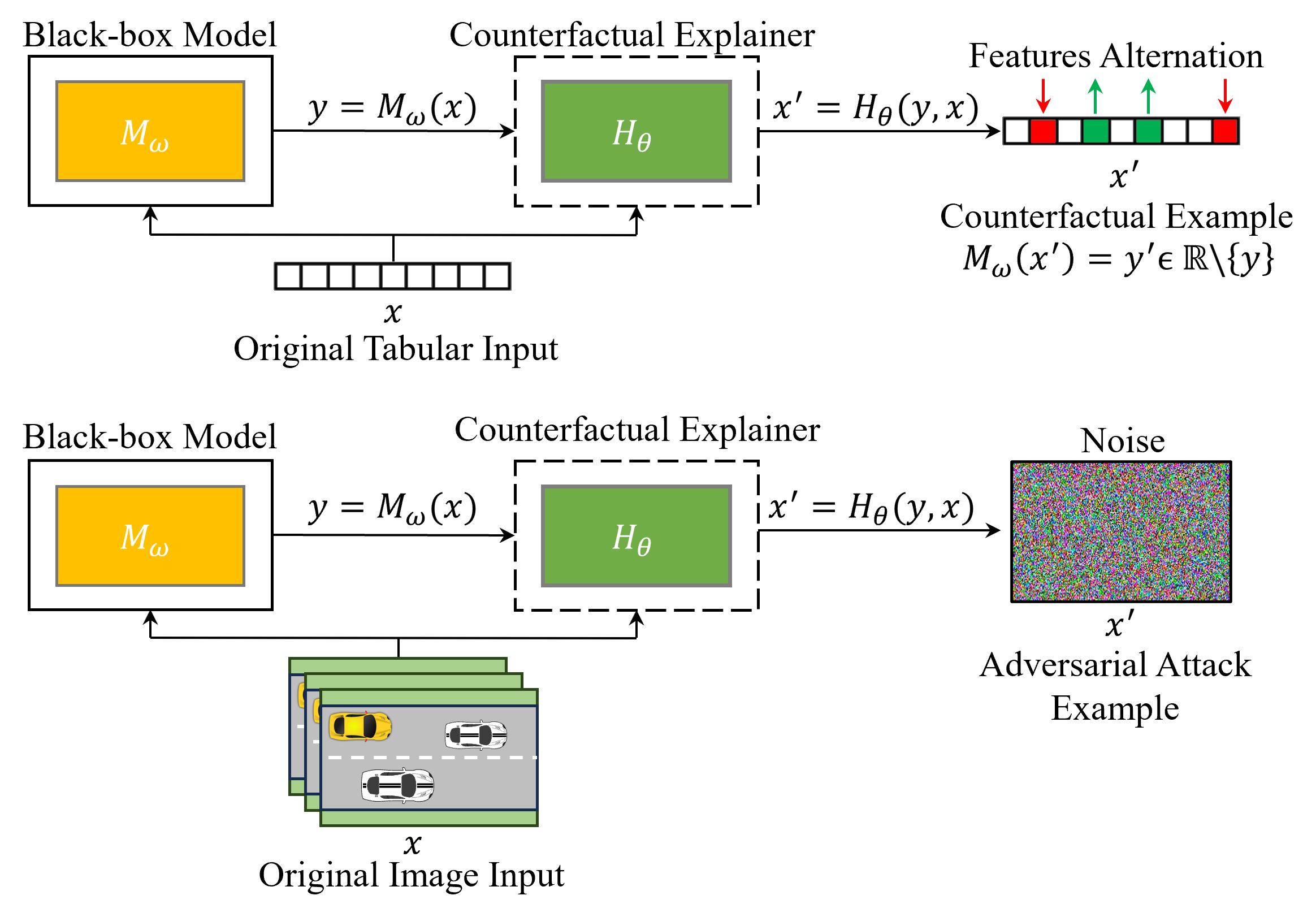}
\caption{\label{fig:CF}Overview of the CF explanation method. Conventional CF methods could be able to generate CFs for (low-dimensional) tabular data; However, for (high-dimensional) image input data, they generate noise (adversarial attack examples). The boxes outlined in solid and dashed lines indicate fixed model parameters and training model parameters, respectively.}
\vspace{-0.5cm}
\end{figure}

\subsection{Counterfactual explanations} 
\label{subsection:counterfactual_explanation}
Counterfactual (CF) explanations, as described in the study in~\cite{wachter2017counterfactual}, attempt to interpret why a model $M_\omega$ classifies/regresses an input $x$ into class/value $y\in {\mathbb{R}}$ rather than a counter class/value $y'\in \mathbb{R}\backslash y$ (see the upper row in Fig. \ref{fig:CF}). Towards this, a CF example called $x'$ that is near $x$ is generated, but the model $M_\omega$ classifies/regresses it as $y'$. The main challenge lies in generating a CF $x'$ in the input-dimension space by calculating the gradient of an assumed loss function between the classes/values $y$ and $y'$ in the output-dimension space. That is why the gradient-based CF explanation method was initially designed for low-dimensional input space problems, such as credit scoring tasks \cite{wachter2017counterfactual}. Applying the same method to high-dimensional input space, such as a high-resolution image, leads to the problem of generating adversarial attacks \cite{shao2014learning,browne2020semantics,goodfellow2014explaining,moosavi2016deepfool,szegedy2013intriguing}, i.e., imperceptible alternations to the input image that deceive the classifier. 

\begin{figure}
\centering
\includegraphics[width=\linewidth]{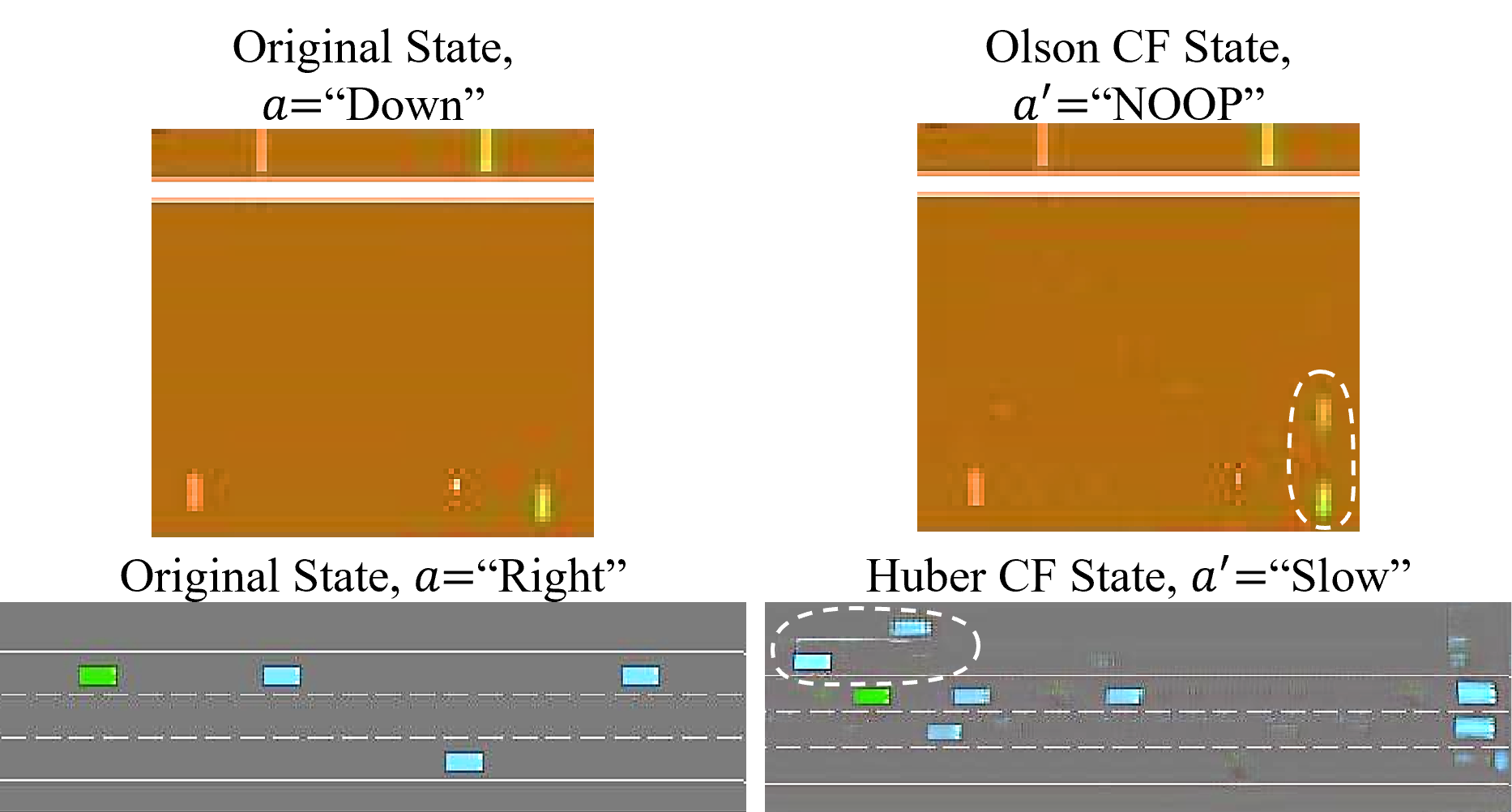}
\caption{\label{fig:GANCF}Generated CF examples for the Pong and Highway environments by the Generative-CF explainer method proposed by Olson~{\it{et. al}}~\cite{olson2021counterfactual} and Huber~{\it{et. al}}~\cite{huber2023ganterfactual}, respectively. For a given initial state (left images), the methods generated implausible CF states on the right (right images). In the Pong environment, the CF state contains two green paddles for the player, and in the Highway environment, two participant vehicles are generated offside of the road, which are out of the rules.}
\vspace{-0.5cm}
\end{figure}

Although the main idea behind adversarial and CF examples are the same, they are applied in two distinct ways. While the input changes in the CF method are meaningful and plausible, they are not noticeable in the adversarial attack and are typically perceived only as additive noise at the output (see the lower row in Fig.\ref{fig:CF}). The first attempts to provide CF explanations for high-dimensional input models were to compare an image $x$ to other \emph{real instances} $x'$ labelled as $y'$ to explain a decision \cite{goyal2019counterfactual, hendricks2018grounding,wang2020scout}, rather than varying the original image's ($x$) features. 
Afterwards, generative model-based CF methods were proposed to shift the model's classification by providing a CF image $x'$ similar to the original image $x$, but slightly different in a few high-level attributes. Progressive Exaggeration (PE)\cite{shrikumar2017learning}, DiVE \cite{rodriguez2021beyond}, STEEX \cite{steex} and Counterfactual State (CFS) \cite{olson2021counterfactual} are remarkable studies that used the generative model to generate CFs. However, as has been emphasised in \cite{olson2021counterfactual}, the generative models cannot guarantee that the generated image is meaningful, as they produce a synthetic image rather than modifying the original image specifications. 
To demonstrate this in practice, we have implemented the two state-of-the-art generative-CF explainers proposed by Olson et al. \cite{olson2021counterfactual} and Huber {\it{et al.}}~\cite{huber2023ganterfactual} to generate CFs for the Atari Pong and Highway environments. Fig. \ref{fig:GANCF} illustrates implausible CF explanations (right frames) generated from the original input images (left frames), generating two paddles for the right player or locating participant vehicles on the offside of the road. 

In our approach, we propose a framework that uses the original data and applies sparse changes through the gradient-based method to ensure that all generated CFs are meaningful and plausible. This is made possible by using policy distillation and transferring the high-dimensional input space to a low-dimensional input space in order to mitigate the problem of adversarial attack. 

\subsection {Policy Distillation}
Some RL methods such as PPO~\cite{schulman2017proximal} and A2C~\cite{mnih2016asynchronous} employ Policy Gradient (PG) techniques that directly optimise the policy by taking the gradient of the expected returns. Another RL method is the Q-Learning algorithm that learns an intermediate quality function, namely the Q-function, for all the state-action pairs $(s, a)$~\cite{mnih2015human}. Finally, there are methods that lie between policy optimisation and Q-learning, such as DDPG~\cite{lillicrap2015continuous}, TD3~\cite{fujimoto2018addressing} and SAC~\cite{haarnoja2018soft}. Due to the high dimensionalities of the state and/or action spaces in complex environments, deep neural networks (DNNs) are used in RL algorithms to map states to values or state-action pairs to Q-values, known as DRL approaches. While advanced end-to-end DRL algorithms, which map high-dimensional states to actions, offer superior performance, their deployment is impeded by the challenge of interpreting their black-box policies. Interpreting such policies is a vital step to establish trust and ensure effective debugging. To address this challenge, we distil the teacher's policy to a shallow neural network with a lower input dimension to interpret the black-box policy while avoiding the adversarial attack illustrated in Fig.~\ref{fig:CF}. Noteworthy, any DRL algorithm can be used in the proposed framework.

\begin{figure}
\centering
\includegraphics[width=\linewidth]{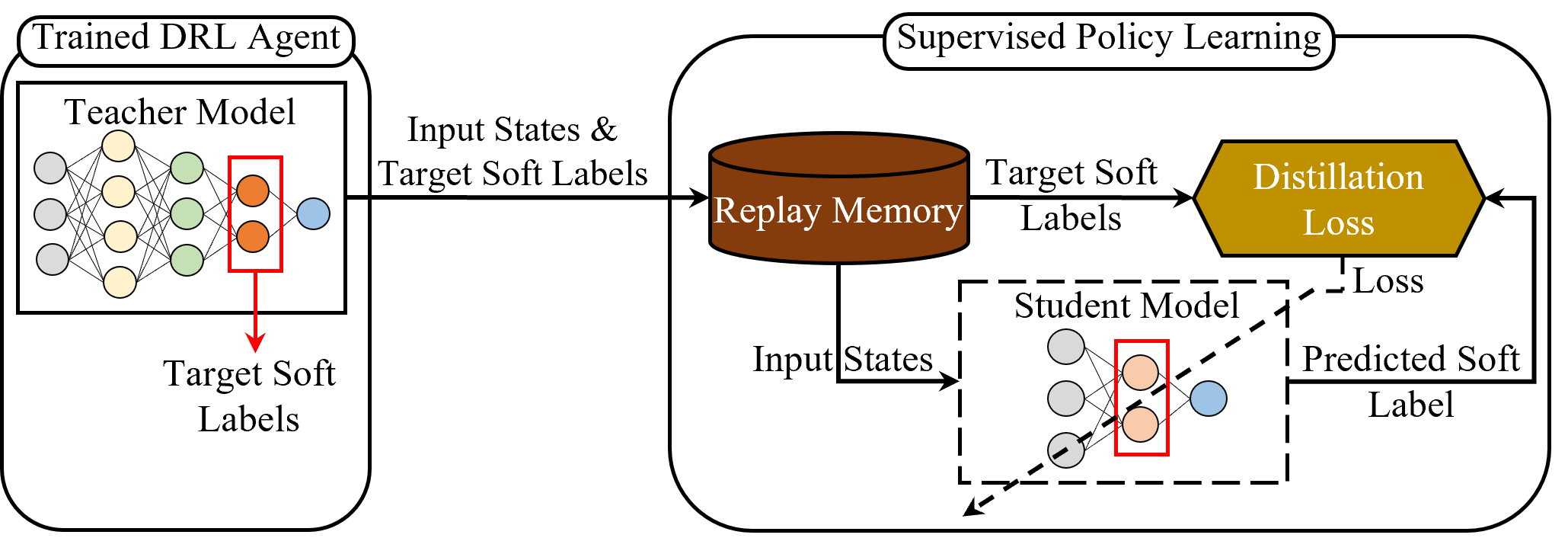}
\caption{\label{fig:PD}Overview of the policy distillation method.  The boxes outlined in solid and dashed lines indicate fixed model parameters and
training model parameters, respectively}
\vspace{-0.4cm}
\end{figure}

In machine learning, Knowledge Distillation (KD) is the transfer of knowledge from a complex model (teacher) to a simpler model (student). It has been proposed in~\cite{ba2014deep,bucil2006model,henelius2014peek}, and in the context of DRL is often called policy distillation\cite{rusu2015policy}. In order to transfer knowledge, the teacher's soft labels (softmax of Q-values in the case of DQN) are passed onto the student network using a KD loss function during supervised learning (see Fig.~\ref{fig:PD}). The Mean Square Error (MSE), the Negative Log likelihood (NLL), and the Kullback-Leibler divergence (KL)~\cite{hinton2015distilling} are the most popular KD loss functions. Also, some studies in the literature use a linear combination of KL divergence and NLL loss functions to shape the KD loss function~\cite{kim2021comparing}. In the literature, the policy distillation paradigm has been used in various applications, including
improved multitask learning~\cite{teh2017distral, traore2019discorl},  deep neural network size compression~\cite{rusu2015policy, schwarz2018progress}, improved learning generalisation \cite{igl2020transient} and policy pruning~\cite{xu2022lrp}. 


\section{Generating CFs Using Policy Distillation}
\label{section: method}
This section describes the proposed framework for generating CF explanations for the actions of a DRL agent using a policy distillation approach implemented into four steps. The aim is to find CFs that can correctly interpret the actions of the trained, black box DRL agent. 

\begin{figure}
\centering
\includegraphics[width=\linewidth]{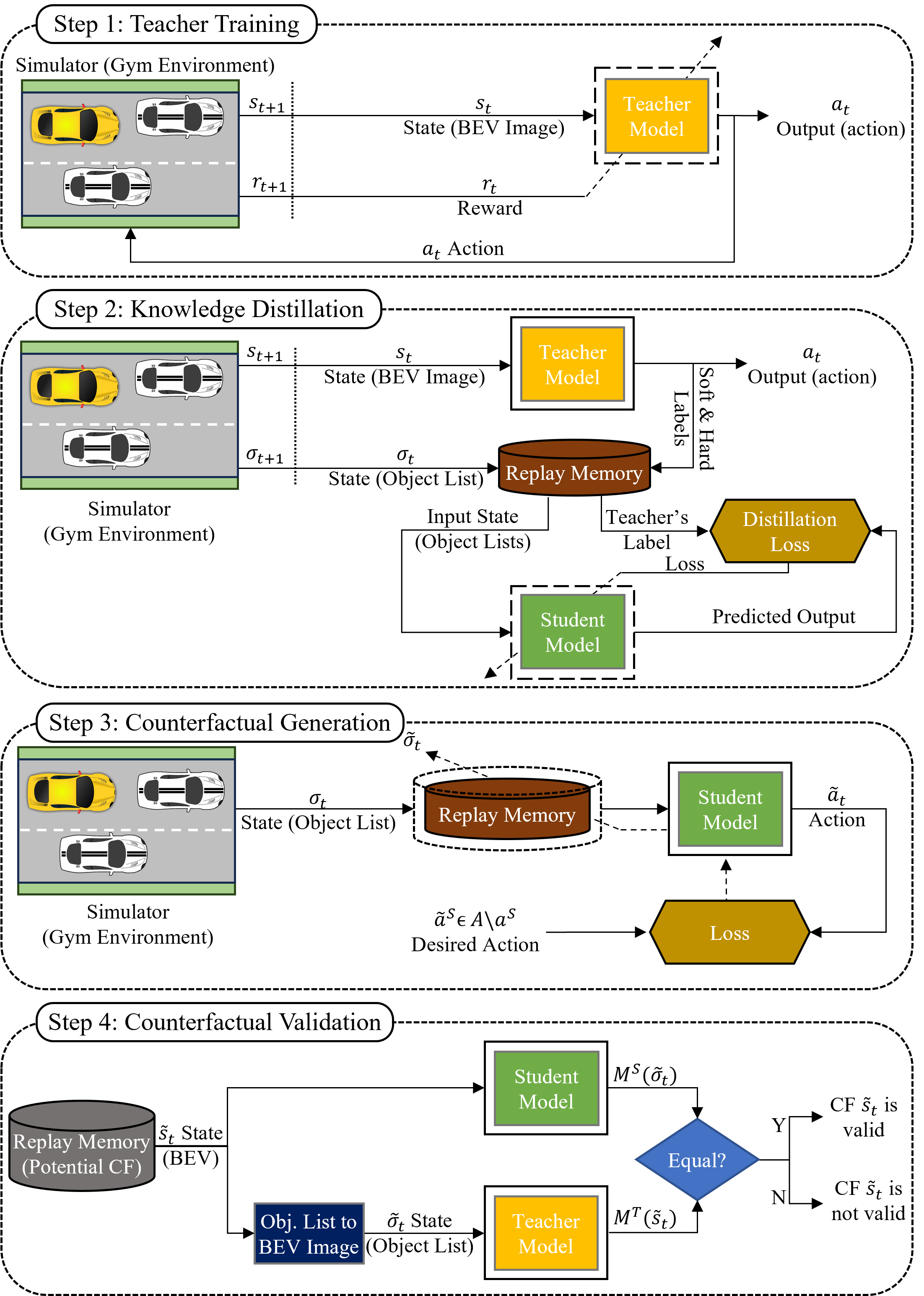}
\caption{\label{fig:method}The proposed CF explainer framework in a highway driving environment. The states $s_t$ and $\sigma_t$ are the states of the environment at time $t$ in the high- and low-dimensional representation, respectively, hence $s_t\in S$ and $\sigma_t \in \Sigma$. Also, $r_t\in R$ denotes the received reward at time $t$ which is used to train the teacher network. The boxes outlined in solid
and dashed lines indicate fixed model parameters and training model parameters, respectively.}
\vspace{-0.4cm}
\end{figure}

The first step (see Step~1 in Fig.~\ref{fig:method} for a block diagram) is referred to as teacher learning, in which the agent learns to act using a high-dimensional input space to describe the state of its environment.  Considering that it is intractable to find the CFs in high-dimensional input space, in the second step (see Step~2 in Fig. \ref{fig:method}), the agent's policy is distilled to a straightforward neural network called the student network that employs low-dimensional inputs. The outcome of the second step is that the student network learns to imitate the teacher's policy within the same environment upon the reception of low-dimensional inputs instead of the teacher's high-dimensional input, e.g., state object lists instead of raw state images. Fortunately, this is possible because most environments provide both high- and low-dimensional states, e.g., in the case of automated driving systems (ADS), a bird's eye view (BEV) image of a highway can be seen as extensive information, while the positions and velocities of vehicles in the scene can be considered as the human-understandable information.

In the third step of the proposed framework (see Step~3 in Fig.~\ref{fig:method}), the potential CFs are generated by calculating gradients of the output (actions) of the student network with respect to the low-dimensional input. In other words, the minimum required changes of the input to obtain the desired system's output are computed. Finally, in the last step (see Step~4 in Fig.~\ref{fig:method} step 4), we evaluate whether the obtained candidate CFs can alter the DRL agent's output (teacher network). In the subsequent paragraphs, each step is detailed in a separate subsection.

\subsection{Teacher Model Training}
In the first step of the policy distillation framework, we train a DRL model to learn how to act within its environment. 
We refer to the trained model as the teacher model $M^T$ since, in the next step, a shallow neural network model (the student model $M^S$) will also be trained to imitate the teacher's policy based on the labels that the teacher's model provides. 

In RL algorithms, an agent learns to act by maximising its expected cumulative future reward obtained through its actions within an environment.
The agent first observes the current state of the environment and the gained reward from its previous action. Based on that it takes another action, and the environment returns its new state and reward to the agent. Through this iterative process, the agent learns an optimised policy, $\pi$, of mapping states onto actions that maximises the cumulative received reward over a time horizon. 

Mathematically, an RL problem is formulated as a Markov Decision Process (MDP) of $(S, A, T, R, \gamma)$, where $S$ and $A$ stand for the states of the environment and the available actions, respectively, $T$ represents the learned policy as a conditional probability distribution function, i.e.~${T: S\times A \times S \rightarrow[0,1]}$ assigns the transition probabilities from the current state $s_t \in S$ to the next state $s_{t+1} \in S$ by taking action $a_t \in A$, and finally, $R$ and $\gamma\in [0, 1]$ are the environment's reward and the discount factor, respectively. The discount factor sets the agent's preference for acquiring long-term rewards ($\gamma$ nearer to unity) over immediate rewards ($\gamma$ nearer to zero). Maximising the expected discounted future reward leads to an estimation of the optimal policy $\pi$. 

\subsection{Student Model Training}
Policy distillation is a technique of policy transferring from a teacher model $M^T$ to a student model $M^S$ in order to learn to imitate the teacher's policy $\pi^T$~\cite{rusu2015policy}. 
To this end, the soft labels of the teacher network (applying a softmax function to the last network layer's weighted sums) are passed to the student model through the distillation loss function. Three loss functions, including the MSE,  $\mathcal L_{MSE}$, the NLL, $\mathcal L_{NLL}$  and the KL divergence, $\mathcal L_{KL}$, are commonly utilised to measure the distillation loss given the hyper-parameters $\theta$ of the student network as follows:
\begin{equation}\label{eq:loss functions}
\begin{split}
&\mathcal L_{MSE}(\mathcal D, \theta)=\sum_{i=1}^{N}\Vert v_i^T - v_i^S \Vert_2^2, \\
&\mathcal L_{NLL} (\mathcal D, \theta)=-\sum_{i=1}^{N}\log P(a_i^S=a_{i,best}^T\vert \sigma_i,\theta),\\
&\mathcal L_{KL} (\mathcal D, \theta)=\sum_{i=1}^{N} \mathrm{softmax}(\frac{v_i^T}{\tau})\ln\frac{\mathrm{softmax}(\frac{v_i^T}{\tau})}{\mathrm{softmax}(v_i^S)}\\
\end{split}
\end{equation}

 With reference to equation~\eqref{eq:loss functions}, the proposed framework provides to the student network an $N$-pair dataset, $\mathcal D$, of low-dimensional object lists (converted from the high-dimensional state offered by the environment) alongside the labels generated by the teacher model. Let us denote by $\sigma_i$ the low-dimensional state associated with the state $s_i$ in the teacher network at the $i$-th time step. Furthermore, we employed $v_i^T$ notation to introduce the soft action of the teacher network, which will be used to train the student model. Therefore, the dataset provided to the student network can be denoted as $\mathcal D=\{(\sigma_i,v_i^T)\}^N_{i=1}$. The parameter $v_i^S$ in equation~\eqref{eq:loss functions} is the soft action generated by the student model, $\tau>0$ is the temperature hyper-parameter and $P(\cdot|\cdot)$ denotes conditional probability distribution function. Finally, the parameter $a_i^S$ is the action taken by the student network and ${a_{i,best}^T=\arg\max_i v_i^T}$ is the optimal action of the teacher network at the same time step. Note that the actions of the teacher and the student networks belong to the same action space, i.e., $a^T\in A$ and $a^S\in A$.

Considering the loss functions in equation~\eqref{eq:loss functions}, it is intuitive to deduce that selecting the MSE loss function maintains the action values $v^S$ generated by the student model close to the teacher's action values $v^T$ by minimising the Euclidean distance between the vectors $v^S$ and $v^T$. The NLL loss function maintains the same indices for the maximum element of the vectors $v^S$ and $v^T$. Finally, the most frequently employed knowledge distillation (KD) loss function is the KL divergence loss between the softened probability distributions of the two models. Intuitively, swinging the hyperparameter $\tau$ to be greater/less than one results in softening/sharpening the models' output probability distribution that leads to the KL divergence loss function attempting to match the models' logits/labels. In this paper, we will conduct a comparison among the aforementioned loss functions in Section~\ref{section:experiments}.   

\subsection{Counterfactual Explanation Generation}
In this step, the proposed framework generates a CF example for the student model that is able to alter its action for a given state. In the next and last step, we will also have to ensure that the provided CF can alter the teacher model's decision. In the student network, $\Sigma \subseteq \mathbb{R}$ represents the input feature space and the actions $A$ as the output space, which can be either continuous, $A \subseteq \mathbb{R}$, or discrete $A \in \{1,\ldots,K\}$. The student model is a  mapping from the input vector state $\sigma\in \Sigma$ to the output action $a^S\in A$, i.e., $M^S_{\theta}:\Sigma \mapsto A$. Given an input state $\sigma$ leading to the action ${a^S=M^S_\theta(\sigma)}$, a CF example is a modified state $\tilde{\sigma}\neq \sigma$ that leads to another action  ${\tilde{a}^S = M^S_\theta(\tilde{\sigma})\neq M^S_\theta(\sigma)}$, where ${\tilde{a}^S \in A \backslash a^S}$. 
In the case of continuous output (action), the validity of CF state $\tilde{\sigma}$ should satisfy the boundary condition of $|{M^S_\theta(\sigma)-M^S_\theta(\tilde{\sigma})}|> \delta$ where $ \delta \in \mathbb{R}\backslash\{0\}$~\cite{chen2022relax}. 

Intuitively, there could exist several CFs $\tilde{\sigma}$ for a given input state $\sigma$. To select among them, various metrics are considered to be maximised or minimised in the literature, including CF and original state distance, sparsity and plausibility of CFs. The most common metric is the distance between the CF and the original state, i.e., distance functions. In this paper, we will employ the Wachter method, which uses the Euclidean distance between the states $\sigma$ and $\tilde{\sigma}$~\cite{wachter2017counterfactual}. Let us denote the selected (optimal) CF by  $\tilde{\sigma}^*$. The Wachter method uses the following loss function to find $\tilde{\sigma}^*$: 
\begin{equation}
\label{eq:argminmax0}
\mathcal L_{CF}(\tilde{\sigma},\lambda;\sigma,a^S,M^S_\theta)=\lambda(M^S_\theta(\tilde{\sigma})-\tilde{a}^S)^2 + \textnormal{dist}(\sigma,\tilde{\sigma})
\end{equation}
with adjustable variables the CF $\tilde{\sigma}$ and the hyperparameter $\lambda$, and fixed inputs the original output $a^S$ and state $\sigma$, and the model $M^S_\theta$. The function $\textnormal{dist}(\sigma,\tilde{\sigma})$ stands for Euclidean distance between the states $\sigma$ and $\tilde{\sigma}$. Since we prefer to apply sparse changes to the original state, we calculate the closest possible CF $\tilde{\sigma}$ to the original state $\sigma$ such that $M^S_\theta(\tilde{\sigma})$ is equal to the new output $\tilde{a}^S$. To this end, the optimal CF $\tilde{\sigma}^*$ could be identified by  minimising the loss function in equation~\eqref{eq:argminmax0} with respect to the hyperparameter $\lambda$. Algorithmically, the $\lambda$ is maximised under an iterative procedure, such that we reach to the optimal CF $\tilde\sigma^*$ associated with the minimum Euclidean distance between the original and CF states, i.e.~${\textnormal{dist}(\sigma,\tilde{\sigma}^*)=\min_{\tilde{\sigma}} \textnormal{dist}(\sigma,\tilde{\sigma})}$. This can be written as: 
\begin{equation}
\label{eq:argminmax}
\tilde{\sigma}^* = 
\arg \min_{\tilde{\sigma}}\max_{\lambda} \lambda(M^S_\theta(\tilde{\sigma})-\tilde{a}^S)^2 + \textnormal{dist}(\sigma,\tilde{\sigma}).
\end{equation}

\subsection{Validation}
In the final step, the proposed framework checks whether the candidate CF can alter the teacher's model output as well as student model's output in the previous step. To do that, the states $\tilde{\sigma}^*$ and $\sigma$ from the student input space (low-dimensional) are first transferred to the teacher input space (high-dimensional). Let us denote them by $\tilde{s}^*$ and $s$, respectively. Next, the framework validates the candidate CF by feeding the teacher model with both states. While, according to the definition of CF explanation, the inequality of the teacher's model output for both the original and CF states indicates the validity of the generated CF, this framework aims to assess the performance of policy distillation by conducting an additional check. Specifically, the validity of a given CF is determined by verifying if both the teacher and student models produce the same action for that CF. The two possible outcomes of the validation process are 
\begin{equation}
\begin{cases}
M^T({s})\neq M^T(\tilde{s}^*) = M^S(\tilde{\sigma}^*) & {validity}=1\\
\textnormal{else} & {validity}=0
\end{cases}
\end{equation}
where ${validity}=1$ indicates that a CF example for the teacher model has been successfully found.


\section{Experiments and Evaluations}
\label{section:experiments}
This section conducts a comprehensive evaluation of the proposed counterfactual (CF)-based explainer method for three black-box DRL agents: DQN\cite{mnih2015human}, A2C\cite{mnih2016asynchronous} and PPO\cite{schulman2017proximal}. These are used in two applications: 1) automated/autonomous driving in highway and roundabout driving scenarios~\cite{highway-env} and 2) Atari Pong game~\cite{1606.01540}. 
Initially, we outline the implementation details and descriptions of the RL environments employed in the experiments. Subsequently, a comparative analysis of the rewards obtained by both the teacher and student models in these environments demonstrates the successful distillation of the teacher's policy into the student model. Finally, we provide a comprehensive quantitative and qualitative comparison of the CFs generated by our proposed PDCF framework against state-of-the-art models, including Olson~{\it{et al.}}~\cite{olson2021counterfactual} and Huber~{\it{et al.}}~\cite{huber2023ganterfactual}. 

\subsection{Implementation Details}
\label{sec:ID}
In our experiments, we employed the DQN, A2C and PPO agents implemented from the stable-baselines3~\cite{stable-baselines3} PyTorch package to serve as the teacher models. For training the student models, we utilised the Adam optimiser with a learning rate of $10^{-3}$.
The KL loss coefficient $\tau$ (refer to Eq.~\eqref{eq:loss functions}) was determined heuristically and set to $0.01$ for the DQN model, and $0.7$ for the PPO and A2C models. To ensure the robustness of the results, we conducted each experiment five times with randomly generated seeds, and subsequently presented the mean values of the extracted metrics. The experiments are executed utilising the PyTorch framework, harnessing the computational power of an NVIDIA RTX-3090 GPU.

\subsection{Automated Driving and Atari Pong Game Environments}
\begin{figure*}
\centering
\includegraphics[width=\linewidth]{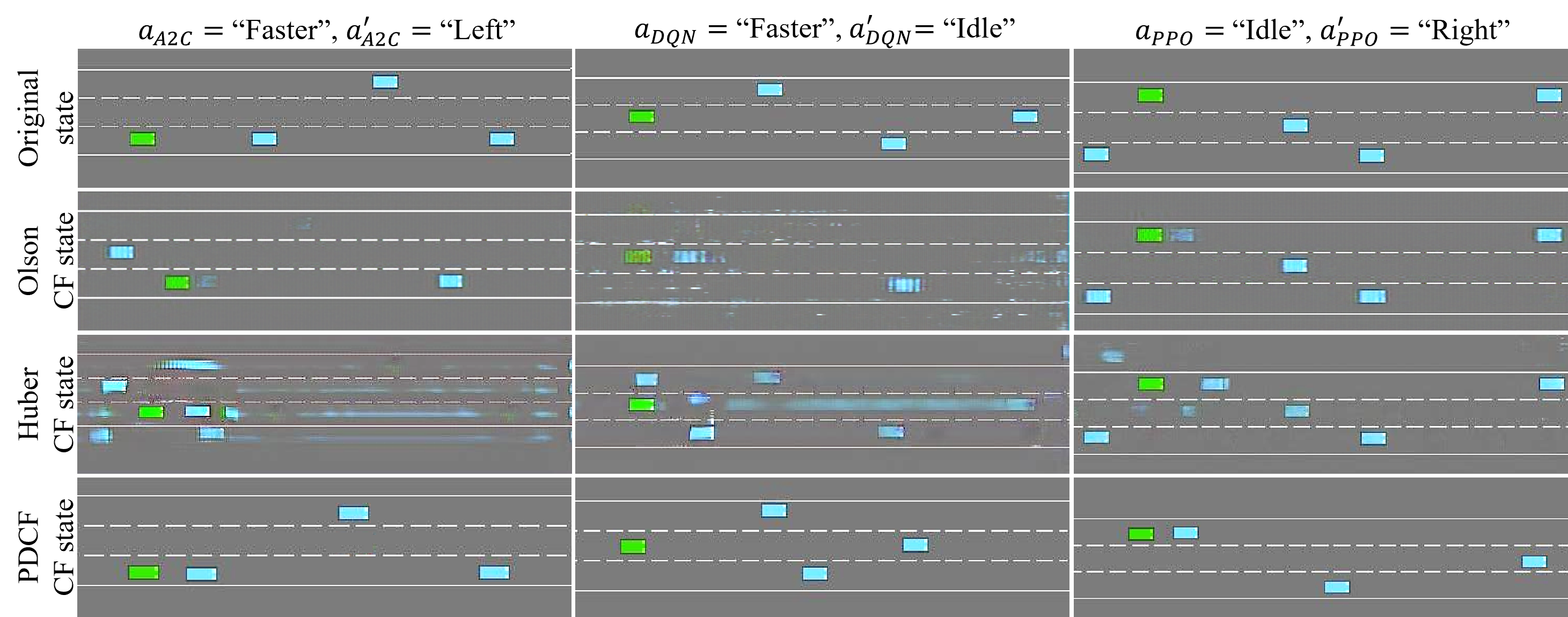}
\caption{\label{fig:highway} 
A visual evaluation of the performance comparison between the SOTA methods and PDCF framework across DRL agents: DQN, PPO, and A2C. The original action is denoted as $a$, and the query action is marked as $a'$. The EV and PVs are depicted with green and blue boxes, respectively.}
\vspace{-0.3cm}
\end{figure*}

\begin{figure}[t!]
\centering
\includegraphics[width=\linewidth]{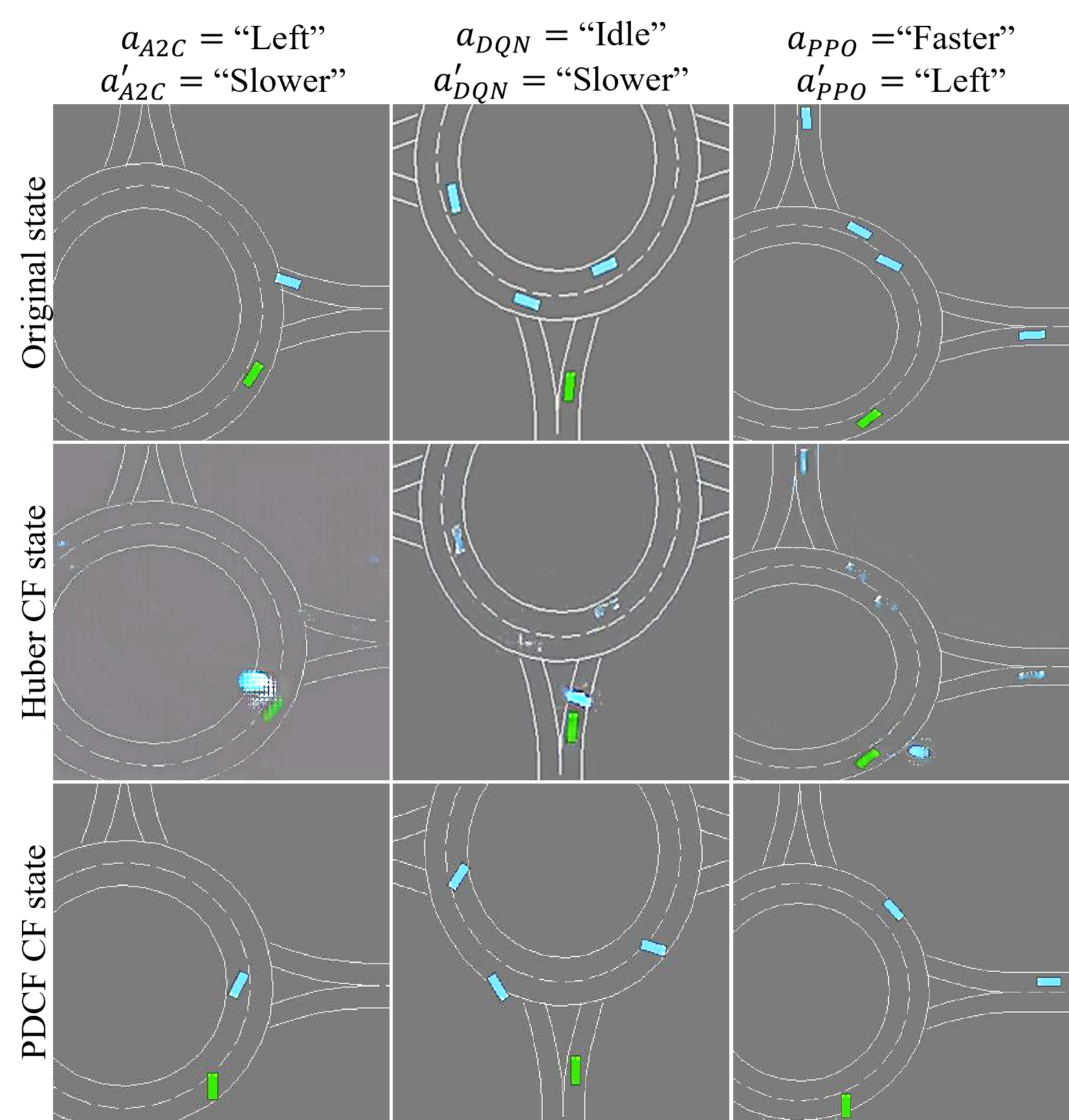}
\caption{\label{fig:roundabout} A visual assessment comparing the performance of SOTA methods with the PDCF framework across three distinct DRL agents: DQN, PPO, and A2C. Notably, the traffic flow in the roundabout is anti-clockwise. The initial action is labelled as $a$, and the target action is denoted by $a'$. The EV and PVs are represented by green and blue boxes, respectively.}
\vspace{-0.5cm}
\end{figure}

The PDCF framework is evaluated in the domains of ADS and the Atari Pong game. The following paragraphs provide descriptions of the  Highway and Roundabout environments, the two driving scenarios implemented as the ADS application, and Atari pong game. 
In the automated driving scenarios, a DRL agent controls the EGO vehicle which is: 
\begin{enumerate}
    \item traveling on a multi-lane highway with left-to-right traffic flow, in the highway environment (Fig.~\ref{fig:highway})
    \item approaching a roundabout with anticlockwise flowing traffic, in the roundabout environment (Fig.~\ref{fig:roundabout}). 
\end{enumerate}
The agent’s objective is to drive at the fastest possible pace while avoiding collisions with Participant Vehicles~(PV). Therefore, the corresponding reward for taking action $a$ at state $s$ can be shaped as follows:
\begin{equation}\label{eq:reward}
R(s,a) = \frac{u-u_{min}}{u_{max}-u_{min}} - \beta \times collision,
\end{equation}
where $R(.,.)$ is the reward function of the environment, $\beta$ is a regulation coefficient, and $u, u_{min}$ and $u_{max}$ are the EGO vehicle's current, minimum and maximum velocities, respectively. In highway and roundabout driving, there are five actions in total comprising ``Left'' lane change, ``Right'' lane change, and keep the lane with ``Faster'', ``Idle'' or ``Slower'' velocity. According to the proposed framework, the DRL teacher model predicts the action by feeding four stacked-state-image observations, while the student network imitates the teacher's policy by receiving object lists, including the vertical/lateral positions and velocities of each vehicle in the environment. 

\begin{figure}
\centering
\includegraphics[width=0.96\linewidth]{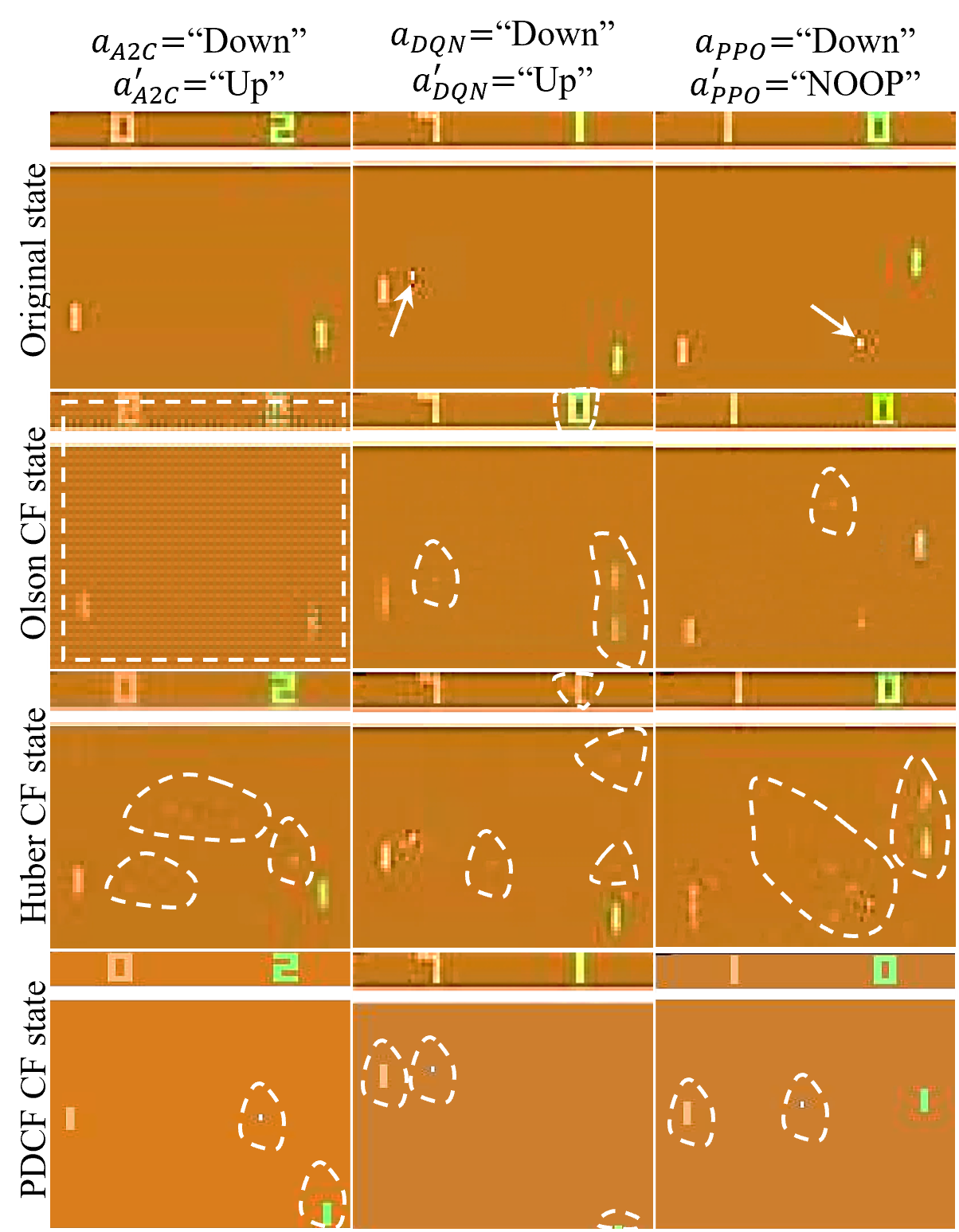}
\vspace{-0.1cm}
\caption{\label{fig:pong}
A visual evaluation comparing the performance of the SOTA methods with PDCF framework across the DRL agents, including DQN, PPO, and A2C. Here, the initial action is denoted as $a$, and the desired action is marked as $a`$. The modified sections of the query states are highlighted by dashed white lines. The ball's movement trajectory is illustrated by white arrows over the specified time period.}
\end{figure}

Using Gym Atari Pong game\cite{1606.01540}, the DRL agent controls the right paddle to play Pong against the built-in opponent of the environment (as shown in Fig.~\ref{fig:pong}). The agent's objective is to deflect the ball away from the right side of the game board (right goal) into the opponent's left goal. The available actions for the agent are ``NOOP'', ``Up'', and ``Down'', which refer to the paddle being moved without change, up, and down, respectively. The agent receives one positive/negative point each time the ball passes the opponent's/agents's goal. Therefore, after 21 rounds of play, the accumulated reward for the agent is $R \in \{-21, \dots ,21\}$. According to the proposed framework, the teacher network predicts the action based on the four stacked-state-image observations, whereas the student network imitates the teacher's policy by receiving only the lateral position of the paddles as well as the vertical and lateral position of the ball as the object lists.

\definecolor{grey}{rgb}{0.9,0.9,0.9}

\begin{table}
\vspace{0.2cm}
\centering
\caption{Teacher and Student model architectures for the DQN, PPO and A2C models.}
\label{architecture}
\begin{adjustbox}{width=\linewidth,center}
\begin{tblr}{
  width = \linewidth,
  column{1} = {c}{1.9cm,l},
  column{2} = {c}{0.95cm,c},
  column{3} = {c}{1.41cm,c},
  column{4} = {c}{1.07cm,c},
  column{5} = {c}{0.0cm,c},
  column{6} = {c}{1.9cm,l},
  column{7} = {c}{1.41cm,c},
  column{8} = {c}{0.6cm,c},
  cell{1}{1} = {c=4}{c},
  cell{1}{6} = {c=3}{c},
  cell{9}{2} = {c=3}{c},
  cell{9}{7} = {c=2}{c},
  row{4,6,8} = {grey},
  hline{1,9-10} = {-}{1pt},
  hline{2} = {1-4,6-8}{},
  hline{3} = {1-4,6-8}{},
  vspan=even,
}
 Teacher Model & & & & & Student Model  & & \\
{\\Layers} & (Filter, Kernel, Stride) & {Output (Highway \& Roundabout)}  & {Output (Atari Pong)} && {\\Layers} & Output (Highway \& Roundabout) & Output (Atari Pong) \\
Input Image  & -            & 128$\times$64$\times$4 &84$\times$84$\times$4 && Input Features & 24  & 16  \\
ReLu(Conv2D) & $($4,8,4$)$  & 31$\times$15$\times$4  & 20$\times$20$\times$4&& ReLu(Linear)   & 256 & 256 \\
ReLu(Conv2D) & $($32,4,2$)$ & 14$\times$6$\times$32  & 9$\times$9$\times$32 && ReLu(Linear)   & 256 & 256 \\
ReLu(Conv2D) & $($64,3,1$)$ & 12$\times$4$\times$64  & 7$\times$7$\times$64 && Linear         & 5   & 12  \\
Relu(Linear) & 3072         & 512                    & 512                  && -              & -   & -   \\
Linear       & 512          & 5                      &3                     && -              & -   & -   \\
Total parameters & 88686246 & highway:23674534,47348042 & pong: && Total parameters &72453&                          
\end{tblr}
\end{adjustbox}
\end{table}

\begin{table}
\caption{Comparison of performance of DQN, PPO, and A2C teachers and their students with identical network architectures for two tasks of highway and roundabout driving.}
\label{tab: policy distilation}
\vspace{-0.1cm}
\begin{adjustbox}{width=\linewidth,center}
\centering
\setlength{\tabcolsep}{0.0pt}
\begin{tblr}{
  column{1-13} = {c},
  column{1} = {l}{0.60cm},
  column{2} = {c}{0.62cm},
  column{3-13} = {c}{0.55cm},
  column{5,9,13} = {c}{0.63cm},
  column{6,10} = {c}{0.0mm},
  cell{1}{1,2} = {r=2}{},
  cell{1}{3,7,11} = {c=3}{},
  cell{3,6,9}{1,3,7,11} = {r=3}{},
  hline{1,12} = {-}{1pt},
  hline{2} = {3-5,7-9,11-13}{},
  hline{3} = {1-13}{},
  row{3-5,9-11}={grey}
}
{DRL\\Model} & {loss\\func} & Highway &&&& Roundabout &&&& Atari Pong & \\
          &       & $T_{s}$ & $S_{s}$&RS\%& & $T_{s}$ & $S_{s}$  &RS\%&& $T_{s}$ & $S_{s}$&RS\% \\
A2C       & KL    & 13.61 & 13.17 & 96.80 && 9.06 & 9.00 & 99.33 && 18.70 &  16.06&85.87 \\
          & NLL   &       & 13.08 & 96.14 &&      & 8.90 & 98.23 &&       &  14.78&79.03 \\
          & MSE   &       & 13.06 & 96.03 &&      & 8.81 & 97.24 &&       &  17.88&95.61 \\
DQN       & KL    & 14.34 & 13.25 & 92.36 && 9.31 & 8.30 & 89.15 && 19.26 &  21.00&109.03 \\
          & NLL   &       & 13.55 & 94.45 &&      & 8.52 & 91.51 &&       &  10.11&52.49 \\
          & MSE   &       & 13.47 & 93.93 &&      & 8.71 & 93.55 &&       &  5.55 & 28.81\\
PPO       & KL    & 12.75 & 12.62 & 98.94 && 9.23 & 8.79 & 95.23 && 19.16 & 17.76 & 92.69\\
          & NLL   &       & 12.72 & 99.76 &&      & 7.63 & 82.66 &&       & 17.59 & 91.80\\
          & MSE   &       & 12.96 & 101.64&&      & 7.90 & 85.59 &&       & 17.76 & 92.69\\          
\end{tblr}
\end{adjustbox}
\vspace{-0.5cm}
\end{table}

\begin{table*}
\centering
\vspace{-0.1cm}
\caption{Evaluation of SAFE-RL in comparison to baseline models. The desired directions of metrics improvement are indicated by arrows ($\downarrow$ low or high $\uparrow$). The most favorable results are highlighted in bold, and the second-bests are underscored. LPIPS, proximity, sparsity, and validity metrics are represented by LPS, Prx, Spr, and Vld, respectively. The methods that were unable to converge are denoted by NC.}
\label{tab: performance}
\vspace{-0.1cm}
\begin{adjustbox}{width=\linewidth,center}
\begin{tblr}{
   column{2-22} = {c},
  cell{1}{1,2} = {r=2}{c},
  cell{1}{3,10,17} = {c=6}{},
  cell{3,6,9}{1} = {r=3}{},
  cell{6}{1} = {r=3}{},
  cell{9}{1} = {r=3}{},
  hline{1,12} = {-}{1pt},
  hline{2} = {3-8,10-15,17-22}{},
  hline{3} = {1-23}{},
  column{1} = {c}{0.65cm},
  column{2} = {c}{1.1cm},
  column{3} = {c}{0.4cm},
  column{3-4,10-11,17-18} = {c}{.63cm},
  column{5,12,19} = {c}{.5cm},
  column{6,13,20} = {c}{.50cm},
  column{7,14,21} = {c}{0.55cm},
  column{8,15,22} = {c}{0.50cm},
  column{9,16} = {c}{0mm},
  row{3-5,9-11} = {grey},
}
{DRL\\Model} & {CF\\Method} & Highway &&&&&&& Roundabout &&&&&&& Pong &&&&& \\
    && $\uparrow$Vld\% & $\downarrow$Spr\% & $\downarrow$Prx &  $\downarrow$FID & $\downarrow$LPS & $\uparrow$IS &
    &   $\uparrow$Vld\% & $\downarrow$Spr\% & $\downarrow$Prx &  $\downarrow$FID & $\downarrow$LPS & $\uparrow$IS &
    &   $\uparrow$Vld\% & $\downarrow$Spr\% & $\downarrow$Prx &  $\downarrow$FID & $\downarrow$LPS & $\uparrow$IS \\
A2C &Olson~\cite{olson2021counterfactual}&\ul{88.38}    &89.28        &\ul{2.413}    &\ul{0.088}    &\ul{0.164}    &\ul{1.112}    &&NC            &NC           &NC            &NC            &NC            &NC            &&\textbf{95.64}&92.76        &5.560         &5.065         &0.417         &\ul{1.230}\\    
    &Huber~\cite{huber2023ganterfactual} &78.05         &\ul{59.25}   &3.636         &1.447         &\textbf{0.130}&\textbf{1.290}&&28.08         &100          &\textbf{0.073}&4.778         &0.362         &\textbf{1.964}&&67.87         &\ul{80.09}   &\ul{1.371}    &\ul{0.266}    &\ul{0.120}    &\textbf{1.390}\\
    &PDCF                                &\textbf{97.38}&\textbf{1.51}&\textbf{1.067}&\textbf{0.002}&0.270         &1.021         &&\textbf{68.00}&\textbf{0.35}&0.253         &\textbf{0.003}&\textbf{0.253}&1.364         &&\ul{76.00}    &\textbf{0.15}&\textbf{0.140}&\textbf{7e-6} &\textbf{0.103}&1.072\\
DQN &Olson~\cite{olson2021counterfactual}&\textbf{99.82}&99.27        &12.99         & 2.049        &0.405         &\textbf{1.486}&&NC            &NC           &NC            &NC            &NC            &NC            &&\textbf{91.81}&95.58        &4.369         &\ul{0.178}    & 0.163        &\ul{1.157}\\
    &Huber~\cite{huber2023ganterfactual} &73.88         &\ul{79.50}   &\ul{2.739}    &\ul{1.353}    &\textbf{0.127}&\ul{1.205}    &&\textbf{81.65}&100          &\textbf{0.014}&3.505         &\textbf{0.150}&\textbf{1.981}&&66.25         &\ul{56.80}   &\ul{1.012}    &0.181         &\textbf{0.087}&\textbf{1.443}\\
    &PDCF                                &\ul{87.38}    &\textbf{1.78}&\textbf{1.255}&\textbf{0.002}&\ul{0.258}    &1.022         &&60.63         &\textbf{0.51}&0.365         &\textbf{7e-4} &0.271         &1.481         &&\ul{89.27}    &\textbf{0.11}&\textbf{0.107}&\textbf{1e-5} &\ul{0.094}    &1.07\\
PPO &Olson~\cite{olson2021counterfactual}&\ul{80.59}    &88.60        &\ul{2.331}    &\ul{0.866}    &\ul{0.159}    &\ul{1.136}    &&NC            & NC          & NC           &  NC          &  NC          &NC            &&71.28         &91.97        &3.978         &\ul{0.114}    &0.164         &\ul{1.338}\\
    &Huber~\cite{huber2023ganterfactual} &75.19         &\ul{73.38}   &2.426         &1.466         &\textbf{0.122}&\textbf{1.262}&&73.14         &100          &\textbf{0.006}&2.150         &\textbf{0.103}&\textbf{1.715}&&\ul{85.04}    &\ul{82.33}   &\ul{1.409}    &0.130         &\ul{0.136}    &\textbf{1.382}\\
    &PDCF                                &\textbf{97.38}&\textbf{1.94}&\textbf{1.369}&\textbf{0.002}&0.266         &1.024         &&\textbf{86.75}&\textbf{0.43}&0.312         &\textbf{7e-3} &0.255         &1.606         &&\textbf{91.77}&\textbf{0.14}&\textbf{0.131}&\textbf{8e-6} &\textbf{0.102}&1.072
\end{tblr}
\end{adjustbox}
\vspace{-0.3cm}
\end{table*}

\begin{table}
\centering
\caption{Evaluating the impact of $\lambda$ on the quality of generated CF states using CF metrics. Desired directions of improvement are indicated by arrows ($\downarrow$ low or high $\uparrow$).  Optimal results are highlighted in bold. Validity, sparsity and proximity metrics are represented by Vld, Spr and Prx, respectively.}
\label{tab: lambda}
\begin{adjustbox}{width=\linewidth,center}
\begin{tblr}{
  row{1-11} = {c},
  cell{1}{1,2} = {r=2}{},
  cell{1}{3,7,11} = {c=3}{},
  cell{3,6,9}{1} = {r=3}{},
  column{2-13}={c},
  column{1}={l}{0.55cm},
  column{2}={c}{0.45cm},
  column{3,7,11}={c}{0.58cm},
  column{4,8,12}={c}{0.58cm},
  column{5,9,13}={c}{0.51cm},
  column{6,10}={c}{0mm},
  hline{3} = {-}{},
  hline{2} = {3-5,7-9,11-13}{},
  hline{1, 12} = {-}{1pt},
  row{6,7,8} = {grey},
}
{DRL\\Model} & $\lambda$ & Highway         &                   &                 && Roundabout      &                   &                 && Pong            &                   &                \\
             &           & $\uparrow$Vld\% & $\downarrow$Spr\% & $\downarrow$Prx && $\uparrow$Vld\% & $\downarrow$Spr\% & $\downarrow$Prx && $\uparrow$Vld\% & $\downarrow$Spr\% & $\downarrow$Prx\\
A2C          & 0.99      & \textbf{97.38}  & 1.51              & 1.067           && \textbf{68.0}   & 0.35              & 0.253           &&\textbf{76.00}   &0.15               &0.140\\
             & 0.5       & 93.75           & 1.33              & 0.946           && 55.88           & 0.30              & 0.220           &&35.15            &0.08               &0.079\\
             & 0.1       & 47.88           & \textbf{1.02}     & \textbf{0.732}  && 26.38           & \textbf{0.23}     & \textbf{0.168}  &&33.43            &\textbf{0.09}      &\textbf{0.081}\\
DQN          & 0.99      & \textbf{87.38}  & 1.78              & 1.255           && \textbf{60.63}  & 0.51              & 0.365           &&\textbf{89.27}   &0.11               &0.107\\
             & 0.5       & 56.38           & 0.92              & 0.659           && 29.63           & 0.17              & 0.130           &&33.91            &0.08               &0.075\\
             & 0.1       & 23.00           & \textbf{0.36}     & \textbf{0.266}  && 20.125          & \textbf{0.03}     & \textbf{0.027}  &&33.91            &\textbf{0.08}      &\textbf{0.075}\\
PPO          & 0.99      & \textbf{97.38}  & 1.94              & 1.369           && \textbf{86.75}  & 0.434             & 0.312           &&\textbf{91.77}   &0.14               &0.131\\
             & 0.5       & 84.00           & 1.43              & 1.021           && 73.50           & 0.27              & 0.200           &&55.52            &0.11               &0.110\\
             & 0.1       & 22.25           & \textbf{0.43}     & \textbf{0.316}  && 30.62           & \textbf{0.15}     & \textbf{0.097}  &&33.54            &\textbf{0.008}     &\textbf{0.078}  
\end{tblr}
\end{adjustbox}
\vspace{-0.6cm}
\end{table}

\subsection{Quantitative Results: Counterfactual Explanation Performance}
\label{subsection:result_CF} 
To start the evaluation process of the CF-based explainer framework, we first train the three teacher-DRL models considered in this study using high-dimensional state images. Following this, the policies of these teachers are distilled into a shallow student model using low-dimensional objective lists as input (translated from the state images), the teacher's soft labels as targets, and employing KL divergence, NLL, and MSE as loss functions. The architectures of both the teacher and student models are detailed in Table~\ref{architecture}, highlighting the considerable reduction in the number of parameters in the student model compared to the teacher model. Table~\ref{tab: policy distilation} compares the efficacy of policy distillation in terms of generalisation performance in three environments, measured by the average rewards earned by the models throughout $10$ trials, each comprising $15$, $10$, $2000$ episodes for the Highway, Roundabout automated/autonomous driving scenarios and the Atari Pong game, respectively. The relative student score percentage ($\frac{S_s}{T_s}\times100$) is reported for ease of comparison with teacher and student model scores. The proximity between teacher and student model scores suggests successful policy distillation. Subsequently, candidate CF explanations are generated for the student model using the Wachter method~\cite{wachter2017counterfactual}~(Eq.~\eqref{eq:argminmax0}). In the final step, to validate the candidate CFs, we verify whether both the teacher and student models yield the same action for a given CF example input, i.e., we check whether $M^T(\tilde{s}^*)=M^S(\tilde{\sigma}^*)$.

To assess the effectiveness of the generated CF explanations, we employ three standard CF metrics: validity, proximity, and sparsity, as presented in Table~\ref{tab: performance}. ``Validity'' indicates the percentage of CF candidates that successfully induce a shift in the model's output towards the desired target action. Intuitively, a higher validity percentage shows that the framework has not only generated more valid CFs, but also has been more successful in distilling the teacher's policy to the student network. ``Proximity'' and ``sparsity'' are metrics that assess the resemblance between the original and CF states. Proximity measures this similarity by computing the average value of the modified pixels, while sparsity evaluates it by counting the mean number of pixels that have been altered. Such metrics provide insights into the extent of changes made to the original state during the CF generation procedure. Intuitively, sparse CFs may lead to more traceable and human-understandable explanations, as they alter fewer original input features, thereby highlighting the most fundamental features attended to by the black-box model. However, there is a trade-off between generating sparse CFs (low in proximity and sparsity metrics) and more valid CFs (high in validity metric), where allowing more perturbed features increases the chances of obtaining a valid CF~\cite{chou2022counterfactuals}. This trade-off can be controlled by hyperparameter $\lambda$ in Equation~\eqref{eq:argminmax0}, examined in Table~\ref{tab: lambda}. Higher values in the regulator parameter prioritise providing more valid CFs, the first term in Equation~\eqref{eq:argminmax0}, while lower values lead to less distance between original states and CF states, the second term in Equation~\eqref{eq:argminmax0}. This phenomenon is evident in Table~\ref{tab: lambda}, where the validity is decreased $75.13\%$ by changing $\lambda$ from 0.99 to 0.1 for the PPO agent in the Highway environment.     

In addition to standard CF metrics, Table~\ref{tab: performance} also compares generative metrics used to assess the realism of the generated CF examples, such as Fréchet Inception Distance~(FID), Learned Perceptual Image Patch Similarity~(LPIPS) and Inception Score~(IS). FID measures the similarity distance between generated and real states based on their feature representations, LPIPS evaluates perceptual differences between real and fake state images using high-level visual features. IS provides a measure of the quality and diversity exhibited by the counterfactual examples. Such metrics offer insights into the visual fidelity and realism of the provided CF examples. Expectantly, generating CF object-list states and re-rendering them into CF image enabled PDCF framework to provide more realistic CF examples, such that in FID, LPIPS and IS metrics the PDCF outperformed the SOA methods, as shown in Table~\ref{tab: performance}.

\begin{figure}
\centering
\includegraphics[width=0.97\linewidth]{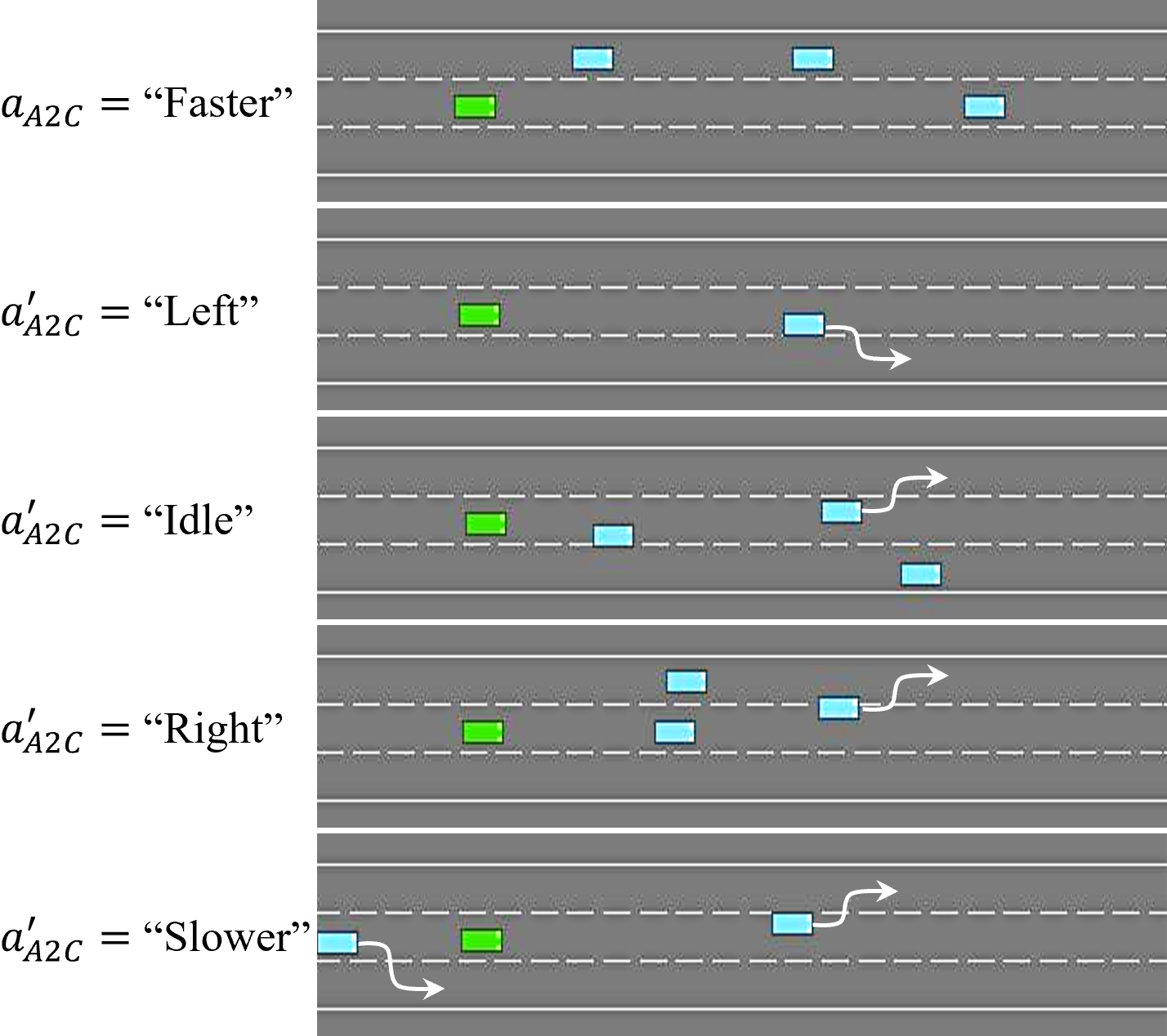}
\vspace{-0.11cm}
\caption{\label{fig:Highway2}
CF explanations generated by the PDCF framework for the A2C agent operating in the Highway environment, covering all possible alternative actions in response to the initial ”Faster” decision.  The lateral velocity direction of PVs is interpreted as their lane change intention, visually represented by arrows.}
\end{figure}

\subsection{Qualitative Results: Counterfactual Explanation Examples}
Fig.~\ref{fig:highway}-~\ref{fig:pong} offer visual comparisons among the PDCF and the SOA CF explainer models, for all mentioned DRL models and environments. The First row frames in Fig.~\ref{fig:highway} present an example illustrations of the query state for the three DRL agents, while the subsequent three rows showcase CF examples generated with baseline methods including Olson~\cite{olson2021counterfactual} and Huber~\cite{huber2023ganterfactual} as well as the proposed PDCF framework. These scenarios offer valuable insights into the effectiveness of the models in generating informative and plausible CF explanations.

For instance, when aiming to alter the decision of the A2C agent from "Faster" to "Left" (Fig.~\ref{fig:highway}, first column), the PDCF framework successfully relocates the front PV (blue box) closer to the EV (green box). This adjustment compels the agent to change its lane to the left, providing a meaningful explanation. In contrast, the method by Huber{\it{et al.}}~\cite{huber2023ganterfactual} produces an unrealistic modification by moving a PV off-road to the left, introducing implausible artifacts. The model proposed by Olson{\it{et al.}}~\cite{olson2021counterfactual} performs better than that of Huber{\it{et al.}}, but it is evident from Figure~\ref{fig:highway} that the DQN agent does not entirely prevent the possibility of a lane change as the PDCF does.

The disparity in quality is also evident in the Roundabout Automated Driving scenarios, as shown in Figure~\ref{fig:roundabout}. When attempting to change the A2C decision from "Left" to "Slower", the PDCF framework successfully repositions the PV in the left lane, altering the lane position and compelling the EV to maintain its current lane while driving at a slower speed. However, in the case of the Huber et al. method, while it attempted to achieve a similar CF, the PV in the generated CF state does not conform to the shape of the other PVs. Additionally, in the case of PPO (Figure~\ref{fig:roundabout}, third column, second row), it produced a PV on the off-side of the road, which is an incorrect representation. Notably, the method proposed by Olson{\it{et al.}}~\cite{olson2021counterfactual} was unable to converge in this environment due to the high-dimensional pixel size of the state images.    

The PDCF method consistently demonstrates high-quality CF explanations, even in the challenging Atari Pong game environment. For example, when considering an initial state classified as "Down" by the PPO model (Figure~\ref{fig:pong}, third column), the PDCF model effectively generates a modified version where a new ball is positioned further away from our paddle, as indicated by the dashed blue lines. This alteration leads to a decision change to "NOOP" (Figure~\ref{fig:pong}, fourth row). In contrast, the state-of-the-art (SOTA) models face difficulties in achieving such distinct adjustments. Olson~{\it{et al.}}'s method yields two paddle balls, failing to properly obscure the original ball. Huber~{\it{et al.}}'s method provides two paddles, multiple balls, and introduces unrelated alterations. The SAFE-RL model consistently introduces sparser changes, underscoring its superior performance in generating substantively meaningful CF explanations when compared to the baseline methods.

\section{Discussion: counterfactuals Explanations for Stakeholders Insight}
\label{sec: Discussion}

This section explores the potential benefits of generating Counterfactual Examples (CFs) in elucidating the rationale behind learned policies for various stakeholders. Existing literature suggests that CF explanations serve several crucial functions: Firstly, they facilitate the debugging of failure cases by providing researchers and engineers with additional insights into corner cases. Secondly, they enhance trustworthiness for end-users by making the behavior and decision-making process of the DNN system transparent. Lastly, they offer justifications to comply with legislation, such as the "right to explanation" regulation outlined in the General Data Protection Regulation (GDPR) applicable across the European Union. In Figure \ref{fig:Highway2}, we showcase the CF explanations generated by the PDCF framework for all potential counter-actions. This demonstration aims to clarify how CF examples can offer comprehensible insights for both end-users and developers. Commencing with the initial state associated with the specified action "Faster" by the A2C agent in the top row, the PDCF framework has generated four distinct CF states for each desired action, involving the following modifications:

\begin{itemize}
\item ``Left'': The front PV is moved closer, indicating an intention for a right lane change.
\item ``Idle'': The PVs in the left and front lanes are brought closer, indicating an intention for a left lane change.
\item ``Right'': The left and front lanes are closed with PVs.
\item ``Slower'': A PV is introduced in the back, indicating an intention for a right lane change, and another PV is generated in front, indicating a left lane change.
\end{itemize}

These clear and informative changes provided by the PDCF framework indicate that the underlying policy of the A2C agent is rational, as it necessitates reasonable adjustments for decision alterations.

In addition to visually justifying the usage of CF explanations to gain insight into DRL models, we offer two straightforward yet informative techniques, namely ``Time-to-Collision Explanation'' and ``Sample-based DRL Agent’s Rule Extraction'' for automated driving systems.  

\subsection{Time-to-Collision Explanation}
\begin{figure}
\centering
\includegraphics[width=\linewidth]{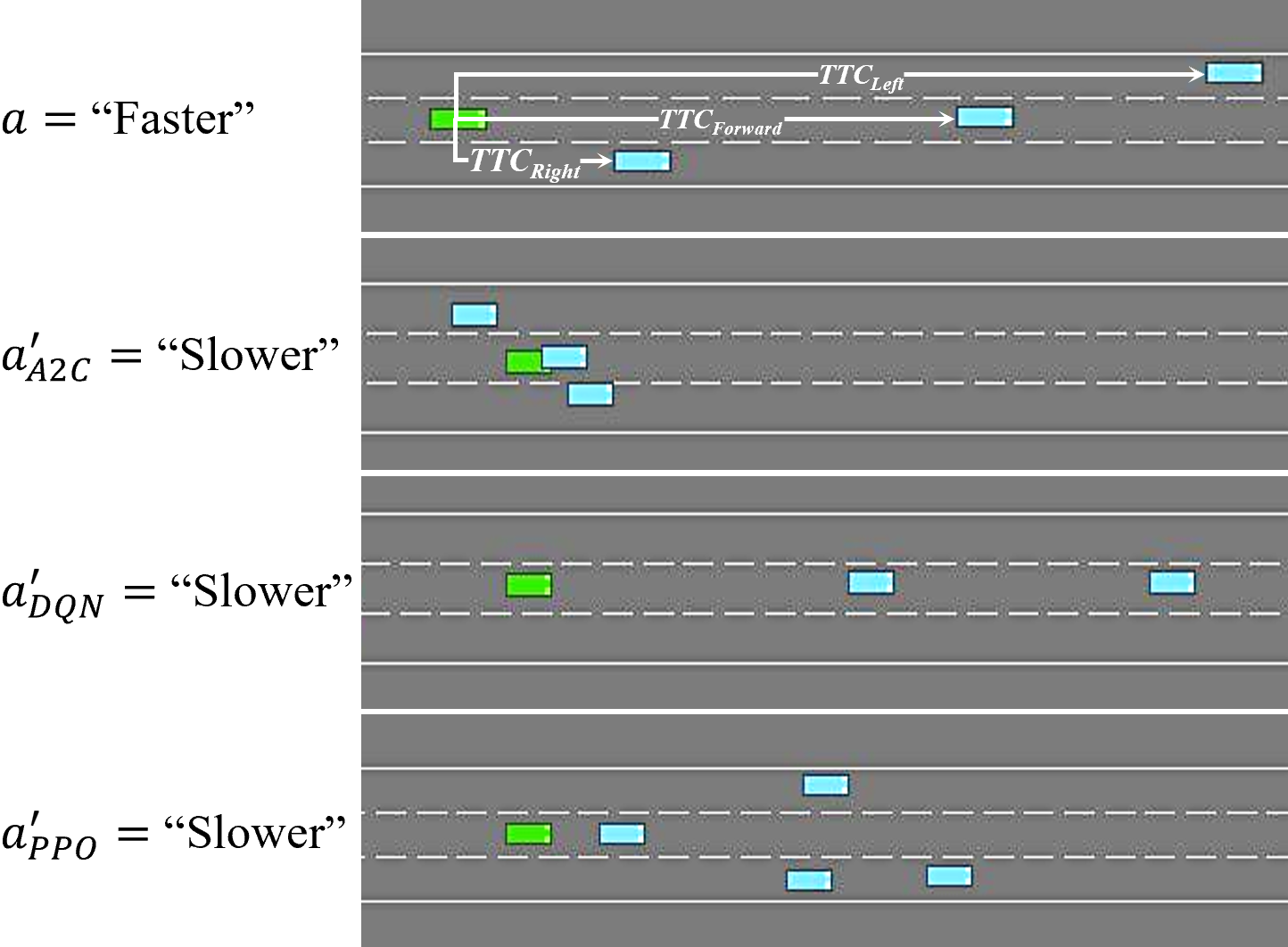}
\caption{\label{fig:TTC} A driving scene for analysing and comparing the EGO vehicle's behavior for two DRL models.}
\end{figure}

Generating multiple CFs can be employed to estimate the decision boundaries of black-box DRL models, enabling engineers to analyse their boundary conditions and interpret their decision-making. 
For instance, consider the top row in Fig.~\ref{fig:TTC}, which illustrates a driving scene marked with the time-to-collision (TTC) values between the EGO Vehicle (EV) and the left, forward, and right Participant Vehicles (PVs). While all DRL models select the ``Faster'' action in this scene, the DQN agent maintains a greater TTC to the leading vehicle ($TTC_{Forward}$) compared to the A2C and PPO agents. This suggests that the DQN agent exercises more caution than the other agents. The result demonstrates that the A2C agent will switch lanes only when the $TTC_{Forward}$ is extremely small. Generating these CFs enables engineers to analyse and, if necessary, tweak their model's behavior. For instance, in this case, the A2C's discount factor $\gamma$ could be adjusted closer to unity to prevent the agent from adopting a myopic policy (maintaining its velocity unchanged until the last moment). As the DRL environment is configurable, engineers could also impose stricter penalties for collisions on the DQN agent.

\subsection{Sample-based DRL Agent's Rule Extraction} 

Employing the generated CFs to construct the decision tree of agent decisions with respect to the TTC can provide a comprehensive estimation of the agent's behavior. Coppens~{\it{et al.}}~\cite{coppens2020synthesising} collected a dataset from a DRL agent operating in an environment to feed a rule mining algorithm, distilling the DRL policy to a rule mining system. Although the extracted rules can provide interpretability to some degree for the end-users, in complex environments, the agent may not necessarily act near its decision boundaries; hence, the rules do not indicate the boundary conditions. Instead, in this paper, we generate CFs, which are naturally near the decision boundary, as explained by Chou~{\it{et al.}}~\cite{chou2022counterfactuals}. In this example, we have gathered pairs of the generated CFs (transferred them to TTC values, which is a more human-understandable feature) and actions and compiled them as a dataset. Subsequently, by fitting a decision tree to this dataset, we extracted if-then rules. 
For instance here we have provided two extracted rules for the PPO agent in the highway driving environment example as follows:
\begin{align*}
{\scriptstyle\text{Rule 1:} \quad} &{\scriptstyle\text{IF } TTC_{\text{left}}  < 7.34 \text{ AND } TTC_{\text{Forward}}  < 5.96 \text{ AND } TTC_{\text{right}}  <  9.64,}\\ & {\scriptstyle\text{THEN Action} = \text{``Right''}} \\
{\scriptstyle\text{Rule 2:}} \quad &{\scriptstyle\text{IF } TTC_{\text{left}} < 10.48 \text{ AND } TTC_{\text{Forward}} < 4.23 \text{ AND } TTC_{\text{right}} < 3.5,}\\ &{\scriptstyle \text{THEN Action} = \text{``Left''}}
\end{align*}
The extracted if-then rules are straightforward and transparent enough for the end-users to comprehend the reasoning underlying the agent's policy. Although the derived rules include a considerable level of simplification, yet they remain a viable source for representing the agent's behaviour. For legal authorities, future work could lead to the formation of standards, compliant with legislation, by collecting all extracted rules and CF explanation images as templates.

\section{Conclusion}
\label{section: conlusion}

This paper developed a framework that can generate human-understandable explanations for the decisions of black-box DRL-based systems, which is a prerequisite for their adoption in safety-critical applications. This is achieved by offering counterfactual (CF) examples, which can capture the essential features in the input data. The key contribution as compared to the state-of-the-art techniques is that the proposed framework can handle high-dimensional input data, e.g., bird's eye view images of motorway traffic, while securing the plausibility of the generated CF explanations. To do that the framework alters human-understandable features such as vehicles' velocities and positions, rather than generating synthetic input images from scratch as with a deep generative approach. In the end, high validity and low sparsity performance are attained allowing us to accurately represent the decision boundary of the DRL models. We carried out several evaluations of the proposed explainer framework to justify, firstly, its ability to outperform alternative approaches from the literature by generating plausible explanations for several DRL methods (Section~\ref{section:experiments}), and secondly, its potential benefits to various stakeholders (Section~\ref{sec: Discussion}).
As a part of future research, it is important to expand the developed framework to include cases where the environment cannot offer both high-dimensional and low-dimensional observation spaces. \cite{chen2018isolating, mathieu2019disentangling}.

\bibliography{ref}
\end{document}